\pdfoutput=1
\documentclass[11pt]{article}

\usepackage[preprint]{acl}

\usepackage{times}
\usepackage{latexsym}

\usepackage[T1]{fontenc}

\usepackage[utf8]{inputenc}

\usepackage{microtype}

\usepackage{inconsolata}

\usepackage{graphicx}

\usepackage{fontawesome5}

\usepackage{booktabs}
\usepackage{multirow}
\usepackage{float}
\usepackage[normalem]{ulem}
\useunder{\uline}{\ul}{}

\usepackage[all]{nowidow}

\title{Efficient Speech Translation through \\Model Compression and Knowledge Distillation}

\author{Yasmin Moslem \vspace{3pt}\\
    ADAPT Centre \\
    School of Computer Science and Statistics \\
    Trinity College Dublin \\
    Dublin, Ireland
  }

\begin{document}
\maketitle
\begin{abstract}
Efficient deployment of large audio-language models for speech translation remains challenging due to their significant computational requirements. In this paper, we address this challenge through our system submissions to the ``Model Compression'' track at the International Conference on Spoken Language Translation (IWSLT 2025). We experiment with a combination of approaches including iterative layer pruning based on layer importance evaluation, low-rank adaptation with 4-bit quantization (QLoRA), and knowledge distillation. In our experiments, we use Qwen2-Audio-7B-Instruct for speech translation into German and Chinese. Our pruned (student) models achieve up to a 50\% reduction in both model parameters and storage footprint, while retaining \mbox{97-100\%} of the translation quality of the in-domain (teacher) models. 
\end{abstract}

\section{Introduction}

Multimodal foundation models have shown powerful capabilities in different tasks, including speech translation. However, these models are often large and computationally intensive, making them impractical to use in real-world settings with limited resources. To enhance the efficiency of these models, researchers have been investigating diverse approaches to model compression that aim to reduce the computational requirements while retaining performance \citep{Gandhi2023-Distil-Whisper,Peng2023-DPHuBERT,Peng2023-I3D,Treviso2023-Efficient-NLP,Wang2023-Pruning}.

Qwen2-Audio \citep{Chu2023-Qwen-Audio, Chu2024-Qwen2-Audio} is a state-of-the-art foundation model that accepts various audio signal inputs and performs audio analysis or direct textual responses to speech instructions. In the IWSLT 2025's Model Compression track \citep{Abdulmumin2025-IWSLT}, the organizers required that all submissions must be derived from the Qwen2-Audio model. The official languages of the task are English-to-German (EN-DE) and English-to-Chinese (EN-ZH).

We experimented with various approaches including efficient fine-tuning using quantized low-rank adapters with 4-bit quantization (QLoRA) \cite{Hu2021-LoRA,Dettmers2023-QLoRA}, iterative layer pruning based on layer importance evaluation \citep{Peer2022-GreedyLayerPruning,Gandhi2023-Distil-Whisper,Sajjad2023-EffectLayerDropping}, and sequence-level knowledge distillation \citep{Kim2016-KnowledgeDistillation,Crego2016-KnowledgeDistillation,Jooste2022-KnowledgeDistillation,Gandhi2023-Distil-Whisper}. Our experiments are mainly based on \textit{Qwen2-Audio-7B-Instruct}.\footnote{\url{https://hf.co/Qwen/Qwen2-Audio-7B-Instruct}} While Section \ref{sec:experiments} elaborates on our experiments, we can summarize our two submissions for German and Chinese as follows:

\begin{itemize}
    \item \textbf{Setup 1:} This model is the outcome of fine-tuning \textit{Qwen2-Audio-7B-Instruct} in two stages: (a) full fine-tuning with the ACL 60/60 dataset, and (b) QLoRA fine-tuning with 4-bit quantization using the ACL 60/60 dataset augmented with data knowledge distillation from the fully fine-tuned model. This process has achieved 40\% compression in terms of both model parameters and storage size. We discuss the details of this model in Section~\ref{sec:qlora}.
    
    \item \textbf{Setup 2:} This model is a pruned version of \textit{Qwen2-Audio-7B-Instruct}, and was created in multiple stages: (a) full fine-tuning of the baseline model with the ACL 60/60 dataset, (b) layer pruning of the decoder into 24 layers, while all 32 encoder layers were kept intact, (c) full fine-tuning of the pruned model, and (d) QLoRA fine-tuning with the ACL 60/60 dataset augmented with data knowledge distillation from the fully fine-tuned model and a portion of the CoVoST2 dataset to restore the quality of the teacher model. This process has achieved 50\% compression in terms of both model parameters and storage size. We discuss the details of this model in Section~\ref{sec:pruning}.
\end{itemize}

\setlength{\tabcolsep}{4pt} % Default value: 6pt

\begin{table*}[htbp]
\centering
\resizebox{\textwidth}{!}{%
\begin{tabular}{@{}lrrrc@{\hskip0.5em}rrrc@{\hskip0.5em}|cc@{}}
\toprule
& \multicolumn{3}{c}{\textbf{EN-DE}} & \multicolumn{4}{c}{\textbf{EN-ZH}} & &
\textbf{Params} & \textbf{Storage} \\ 
\cmidrule(rl){2-4} \cmidrule(rl){6-8} \addlinespace[3pt]
\textbf{Model}                    & \textbf{BLEU}  & \textbf{chrF++} & \textbf{COMET} & & \textbf{BLEU}  & \textbf{chrF}  & \textbf{COMET} & &
\multicolumn{1}{c}{(B)} & \multicolumn{1}{c}{(GB)}
\\ \midrule

Baseline            & 22.96      & 49.88     & 8.61     & & 38.73   & 32.53 & 17.38 & & \multirow{2}{*}{8.40} & \multirow{2}{*}{16.79} \\
+ Full Fine-tuning  & 39.28      & 65.27     & 56.32    & & 58.54   & 52.54 & 65.97 & &  \\ \midrule

+ QLoRA Fine-tuning & 41.52      & 66.25     & 53.47    & & 57.65   & 51.08 & 65.77 & & \multirow{2}{*}{\textbf{4.95}} & \multirow{2}{*}{\textbf{9.64}} \\

% [full] 
+ Knowledge Distillation & \textbf{43.25} & \textbf{68.02} & \textbf{59.36} && \textbf{59.60} & \textbf{53.94} & \textbf{68.42} & & \\

% + Knowledge Distillation [+lora] & \textbf{43.53} & \textbf{68.12}  & \textbf{57.96} & & \textbf{60.92} & \textbf{54.35} & \textbf{66.22}          \\ 
\bottomrule
\end{tabular}
}
\caption{Evaluation of the experiment that employs QLoRA with 4-bit quantization and augments the authentic data with knowledge distillation. This approach reduces the model size by more than 40\% in terms of both the number of parameters (params) and storage footprint, while achieving the best translation performance for both English-to-German (EN-DE) and English-to-Chinese (EN-ZH) language pairs.}
\label{tab:qlora-kd}
\end{table*}

\section{Data}
\label{sec:data}

% Constrained Data
\subsection{In-Domain Data}

The ACL 60/60 dataset\footnote{\url{https://hf.co/datasets/ymoslem/acl-6060}} \citep{Salesky2023-ACL6060} is used in all of our experiments as the in-domain data.
ACL 60/60 contains multilingual translation of ACL 2022 technical presentations into 10 target languages. The dataset consists of two splits, “dev” and “eval”, which together comprise 884 utterances. We merged the two splits, and randomly sampled 100 utterances for testing, which left us with 784 utterances for training. For test data sampling, we used the \textit{train\_test\_split} method from the \textit{datasets}\footnote{\url{https://github.com/huggingface/datasets}} library \citep{Lhoest2021-Datasets}, setting the random seed option to zero. The ACL 60/60 dataset was required for ``constrained'' submissions to the IWSLT's ``Model Compression'' track.

% Unconstrained Data
\subsection{Out-of-Domain Data}

After layer pruning of the \textit{Qwen2-Audio-7B-Instruct} model (cf.~Section~\ref{sec:experiments}), we needed to use more training data to restore the translation quality of the unpruned model. In addition to knowledge distillation data from the teacher model, we used a portion of the CoVoST2 dataset \citep{Wang2021-CoVoST2} (cf.~Section~\ref{sec:pruning}). CoVoST2 is a large-scale multilingual speech-to-text translation corpus, covering translations from English into 15 languages, including German and Chinese.

\setlength{\tabcolsep}{4pt} % Default value: 6pt
\begin{table*}[htbp]

\centering
\resizebox{\textwidth}{!}{%
\begin{tabular}{@{}llrrrc@{\hskip0.5em}rrrc@{\hskip0.5em}|cc@{}}

\toprule
 & & \multicolumn{3}{c}{\textbf{EN-DE}} & & \multicolumn{3}{c}{\textbf{EN-ZH}} & & \textbf{Params} & \textbf{Storage} \\
 \cmidrule(rl){3-5} \cmidrule(rl){7-9} 
 %\addlinespace[3pt]
\textbf{Model} & \textbf{Data} & \textbf{BLEU} & \textbf{chrF++} & \textbf{COMET} & &
\textbf{BLEU} & \textbf{chrF} & \textbf{COMET} & &
\multicolumn{1}{c}{(B)} & \multicolumn{1}{c}{(GB)} \\ \midrule

Baseline  & - &  22.96 & 49.88 & 8.61 & & 38.73 & 32.53 & 17.38 & & 
\multirow{2}{*}{8.40} & \multirow{2}{*}{16.79} \\
+ Full FT & ACL  & 39.28 & 65.27 & 56.32 & & 58.54 & 52.54 & 65.97 & &  &  \\
\midrule

+ Pruning & ACL & 10.78 & 39.58 & -44.4 & & 42.52 & 36.92 & 39.42 & & 
\multirow{3}{*}{6.78} & \multirow{3}{*}{13.55} \\
+ FT & ACL & 32.16 & 60.39 & 39.08 & & 53.05 & 47.23 & 56.72 &  &  &  \\
& + KD &  33.44 & 60.91 & 39.23 & & 53.41 & 48.20 & 54.94 & &  &  \\

\midrule

\multirow{1}{*}{+ QLoRA} & + CV
& 39.59 & 65.14 & 59.21 & & 56.52 & 50.74 & 64.34 & & 
\multirow{1}{*}{\textbf{4.12}} & \multirow{1}{*}{\textbf{{\hskip0.5em}8.65}} \\

% 50k
% \multirow{1}{*}{+ QLoRA} & + CV
% & 38.47 & 64.09 & 54.56 & & 55.95 & 50.11 & 64.50 & & 
% \multirow{1}{*}{\textbf{4.12}} & \multirow{1}{*}{\textbf{{\hskip0.5em}8.65}} \\

\bottomrule
\end{tabular}
}

\caption{Evaluation of the iterative layer pruning experiment. We started by full-parameter fine-tuning (Full FT) of the baseline model \textit{Qwen2-Audio-7B-Instruct} on the in-domain dataset ACL 60/60. Pruning 8 layers of the decoder of the model achieved approx. 20\% reduction in the model size; however, it affected the quality of the model. Hence, we fine-tuned the pruned model again on the in-domain dataset to restore as much as possible of the quality of the fully fine-tuned model. Finally, we fine-tuned the resulting model with low-rank adaptation after quantizing it into the 4-bit precision (QLoRA) on a mix of the in-domain data, knowledge distillation data (KD) and out-of-domain data, namely the CoVoST2 (CV) dataset. The whole process of pruning followed by QLoRA fine-tuning with 4-bit quantization has resulted in approx. 50\% reduction in the model size, while retaining 97\% and 100\% of the translation quality for Chinese and German, respectively, compared to the teacher model.}
\label{tab:layer-pruning}
\end{table*}

% WITHOUT torch_dtype=torch.bfloat16,
% pruning layer importance was done _with_ torch_dtype=torch.bfloat16,

\section{Methodologies}
\label{sec:experiments}

We experiment with diverse methods to compress \textit{Qwen2-Audio-7B-Instruct} while maintaining the translation quality. This section covers our two main experimental setups, while Section~\ref{sec:ablation-study} discusses several ablation studies and elaborates on intermediate experiments.

Our first experimental setup (cf.~Section~\ref{sec:qlora}) employs QLoRA fine-tuning with 4-bit quantization, while the other experimental setup (cf.~Section~\ref{sec:pruning}) conducts layer importance evaluation and applies iterative layer pruning of the model, followed by QLoRA fine-tuning. Both setups use knowledge distillation to recover the translation quality of the in-domain teacher model.

\subsection{Full-Parameter Fine-tuning}
\label{sec:full-fine-tuning}

In all of our experiments, we start by full-parameter fine-tuning of the \textit{Qwen2-Audio-7B-Instruct} model on the ACL 60/60 dataset. This step is essential to ensure the foundation model is familiar with the downstream task and in-domain data. In particular, we train the baseline model on the in-domain data for 3 epochs, using a batch size of 4, learning rate of 1e-5, weight decay of 0.001, and no warm-up steps. The model is initially loaded in \textit{bfloat16} data type. As shown in Table~\ref{tab:qlora-kd} and Table~\ref{tab:layer-pruning}, the fully fine-tuned model clearly outperforms the baseline by an average of 48 COMET points for translation into German and Chinese. Hence, this fully fine-tuned model is used as a foundation for both our experiments in Section~\ref{sec:qlora} and Section \ref{sec:pruning}. It is also used as a ``teacher'' for knowledge distillation into the compressed ``student'' models. We conduct the training on one H200 SXM GPU, using the Transformers framework\footnote{\url{https://github.com/huggingface/transformers}} \citep{Wolf2020-Transformers}.

\subsection{QLoRA with 4-bit Quantization}
\label{sec:qlora}

This experimental setup employs quantization, accompanied by efficient fine-tuning, and knowledge distillation techniques. We start with full-parameter fine-tuning of \textit{Qwen2-Audio-7B-Instruct} (cf.~Section \ref{sec:full-fine-tuning}). Afterwards, the fine-tuned model is quantized into the 4-bit precision and then fine-tuned with low-rank adapters (QLoRA) \citep{Hu2021-LoRA,Dettmers2023-QLoRA}. Moreover, we use the fully fine-tuned model as a ``teacher'' in the knowledge distillation process.

\paragraph{Knowledge distillation:} To restore the quality of the in-domain fully fine-tuned teacher models (cf.~Section \ref{sec:full-fine-tuning}), sequence-level knowledge distillation is applied \citep{Kim2016-KnowledgeDistillation,Gandhi2023-Distil-Whisper}. In other words, we translate the ACL 60/60 training data with the ``teacher'' model. The training data is then augmented with the knowledge distillation data, and duplicates are filtered out. As a result, the augmented data after knowledge distillation comprises 1,568 segments for German and 1,069 segments for Chinese. Finally, ``student'' models are fine-tuned with QLoRA on the augmented data.

\paragraph{QLoRA fine-tuning:} First, we enable 4-bit quantization through \textit{BitsAndBytes} \citep{Dettmers2023-QLoRA} while loading the in-domain fully fine-tuned model. In the configuration of \textit{BitsAndBytes}, we set the quantization type to ``nf4'', and use the ``double\_quant'' option, where the quantization constants from the first quantization are quantized again. For LoRA configuration, we set the rank to 64, alpha to 128, and dropout to 0. We target all linear modules. Overall, this configuration results in 2.41\% trainable parameters of the model. Moreover, we enable Rank-Stabilized LoRA (rsLoRA) \citep{Kalajdzievski2023-rsLoRA}. We train the model for 4 epochs, using a batch size of 4, and a learning rate of 1e-5.

As Table \ref{tab:qlora-kd} illustrates, the combination of QLoRA with 4-bit quantization and knowledge distillation has achieved the highest translation performance into both German and Chinese across all evaluation metrics while reducing the model size in terms of both the number of parameters and storage requirements by more than 40\% compared to the baseline model.\footnote{For consistency, our storage footprint calculations are based only on the size of model files (\textit{*.safetensors}), including the adapter of QLoRA models. We exclude the tokenizer and configuration files, as their contribution to the overall size is relatively small ($\sim$~13~MB) and they are unaffected by our optimization methods. All sizes are computed using the decimal definition of gigabytes, where \(1~\mathrm{GB} = 1000^3\) bytes.}

\subsection{Iterative Layer Pruning}
\label{sec:pruning}

In this experimental setup, we apply iterative layer pruning to the fully fine-tuned ``teacher'' model (cf.~Section \ref{sec:full-fine-tuning}). This approach incrementally identifies and removes layers with minimal contribution to translation quality, one layer at a time. The pruned model resulting from this process is then fine-tuned on both the ACL 60/60 training dataset and knowledge distillation data from the teacher model. Finally, the model is further fine-tuned with QLoRA, leveraging 4-bit quantization for efficient low-rank adaptation. Pruning 8 layers achieves a 20\% reduction in model size, which increases to 50\% when combined with 4-bit quantization. Fine-tuning the pruned model restores between 97\% and 100\% of the teacher model's translation quality for Chinese and German, respectively. The following points elaborate on the process.

\paragraph{Layer importance evaluation:} We conduct layer importance evaluation by measuring translation performance without each layer. In this greedy layer pruning approach \citep{Peer2022-GreedyLayerPruning,Rostami2024-CULL-MT}, to prune \(n + 1\) layers, only a single optimal layer to prune must be added to the already known solution for pruning $n$ layers. After identifying and removing the least critical layer, we repeat the layer importance evaluation on the remaining layers until reaching our $n$ pruning target. We observe that while removing certain layers of the model (e.g. the first or last layer) substantially degrades translation performance, others result in minimal performance drops. When experimenting with using either the chrF/chrF++ or COMET metric for layer importance evaluation, the models pruned based on chrF/chrF++ outperform those pruned based on COMET.

\paragraph{Layer pruning:} We iteratively prune one decoder layer at a time, selecting the layer whose removal has the least negative impact on translation quality, measured by chrF/chrF++ scores. At each iteration, we evaluate the translation performance of the pruned model on the test split of the ACL 60/60 dataset, after removing each candidate layer. The layer whose removal yields the best performance is eventually pruned. This process continues until a predefined number of layers (8 in the main experiments) have been removed. By iteratively removing the least important layers, this performance-guided method produces a more compact model that can be fine-tuned further to recover the translation quality of the teacher model. We observe that the performance of the English-to-German model is more impacted by pruning than the English-to-Chinese model, which might be attributed to the pre-training process (cf.~Table \ref{tab:layer-pruning}).

\paragraph{Knowledge distillation:} Knowledge distillation is the process of transferring knowledge from a large model (teacher) to a smaller one (student). In our case, the teacher is the fully fine-tuned model, and the student is the model resulting from iterative pruning. We translate the in-domain data with the teacher model to augment the authentic data. As the process can result in duplicate translations, we remove duplicate segments from the training data. This step is similar to what we did in the first experimental setup (cf.~Section \ref{sec:qlora}).

\paragraph{Fine-tuning:} The pruning step is followed by fine-tuning the pruned model for 4 epochs using the in-domain ACL 60/60 dataset augmented with the knowledge distillation data \citep{Kim2023-SOLAR10,Gandhi2023-Distil-Whisper}. This step recovers most of the translation quality of the teacher model.

\paragraph{QLoRA fine-tuning:} This step serves two purposes, improving both the compression level and translation performance of the pruned model. After quantizing the model resulting from the previous step, low rank adaptation is used for fine-tuning it further. The training data consists of the ACL 60/60 dataset augmented with the knowledge distillation data from the fully fine-tuned ``teacher'' model (oversampled by a factor of 10), as well as a portion of the CoVoST2 dataset (100k utterances). We fine-tune the model for 1 epoch, using a batch size of 8 (to make use of the computing resources, since the model is much smaller now), and a learning rate of 1e-5.

This whole process of iterative layer pruning followed by fine-tuning has achieved up to 50\% compression in terms of both model parameters and storage size.\footnote{To achieve these compression gains from layer pruning, the Qwen2-Audio model must initially be loaded in \textit{bfloat16} data type. For other tasks like fine-tuning and inference, \textit{bfloat16} precision is not required, although it may be necessary when computing resources are limited, potentially at the cost of reduced quality.} Moreover, the quality degradation caused by pruning has been mitigated through multi-stage fine-tuning on diverse data. As demonstrated by Table \ref{tab:layer-pruning}, by the end of the process, the pruned model could recover most of the translation quality of the fully fine-tuned teacher model (97\% for Chinese and 100\% for German).

Our ablation study (cf.~Section \ref{sec:ablation-study}) demonstrates that iterative layer pruning considerably outperforms fixed middle-layer pruning (cf.~Section \ref{sec:study-middle-pruning}). Moreover, it clarifies that pruning exclusively decoder layers yields better performance than pruning both encoder and decoder layers (cf.~Section \ref{sec:study-enc-dec-pruning}). While our main experiments in this section prune only 8 layers, resulting in a model with 24 decoder layers and 32 encoder layers, the ablation study investigates pruning up to 16 layers (cf.~Section~\ref{sec:study-more-layers}).

\begin{table*}[!ht]
\centering
\begin{tabular}{@{}lllllll@{}}
\toprule
\textbf{Encoder} & \textbf{Decoder} & \textbf{BLEU} & \textbf{chrF++} & \textbf{COMET} & \textbf{Params} & \textbf{Storage} \\ \midrule
24 ↓ & 24 ↓ & 26.44 & 54.67 & 13.90 & 6.62 B & 13.24 GB \\
32 $=$ & 24 ↓ & \textbf{30.81} & \textbf{58.45} & \textbf{31.95} & 6.78 B & 13.55 GB \\ \bottomrule
\end{tabular}
\caption{Comparison of layer pruning of both the encoder and decoder with layer pruning of the decoder only. Both models are fine-tuned before and after layer pruning on the EN-DE ACL 60/60 dataset. This experiment uses middle layer pruning. The model that prunes the layers of only the decoder outperforms the model that prunes both the encoder and decoder, although the former has a slightly higher number of parameters and storage size.}
\label{tab:study-enc-dec}
\end{table*}

\section{Ablation Study}
\label{sec:ablation-study}

This section elaborates on some intermediate experiments that led us to the final models in Section \ref{sec:qlora} and Section \ref{sec:pruning}. These experiments include comparing the performance of the Qwen2-Audio base model with Qwen2-Audio-Instruct (cf.~Section \ref{sec:study-base}), comparing encoder-decoder pruning with decoder-only pruning (cf.~Section \ref{sec:study-enc-dec-pruning}), comparing ``iterative'' layer pruning with fixed middle-layer pruning (cf.~Section \ref{sec:study-middle-pruning}), iterative pruning of up to 16 layers (cf.~Section \ref{sec:study-more-layers}), fine-tuning before and after pruning (cf.~Section \ref{sec:study-fine-tuning}), and using different sizes of out-of-domain data to fine-tune the pruned models (cf.~Section \ref{sec:study-data-size}).

\subsection{Qwen2-Audio Base vs Instruct}
\label{sec:study-base}

We experimented with both \textit{Qwen2-Audio-7B} and \textit{Qwen2-Audio-7B-Instruct} for English-to-German speech translation. As Table \ref{tab:study-base} the “Instruct” model outperforms its base version. Hence, we use \textit{Qwen2-Audio-7B-Instruct} in all of our experiments throughout the paper.

We follow the prompt requirements of the \textit{Qwen2-Audio} models. For the base model, Qwen2-Audio-7B, the prompt is as follows:

\begin{footnotesize}
\begin{itemize}
 \item [] "<|audio\_bos|><|AUDIO|><|audio\_eos|>Translate the English speech into \{language\}:"
\end{itemize}
\end{footnotesize}

For the instruction-following model, \textit{Qwen2-Audio-7B-Insruct}, the prompt is as follows:

\begin{footnotesize}
\begin{itemize}
    \setlength{\topsep}{0pt}
    \setlength{\itemsep}{0pt}
    \item[] [
    \item[] \hspace{1em}\{"role": "system", "content": "You are a professional translator."\},
    \item[] \hspace{1em}\{"role": "user", "content": [
    \item[] \hspace{2em}\{"type": "audio", "audio\_url": audio\_path\},
    \item[] \hspace{2em}\{"type": "text", "text": "Translate the English speech into \{language\}:"\},
    \item[] \hspace{1em}]\},
    \item[] ]
\end{itemize}
\end{footnotesize}

\begin{table}[htbp]
\centering
\begin{tabular}{@{}llrrr@{}}
\toprule
\textbf{Language} & \textbf{Model} & \textbf{BLEU} & \textbf{chrF/++} & \textbf{COMET} \\ \midrule
\multirow{2}{*}{\textbf{EN-DE}} & base & 6.15 & 32.10 & -58.34 \\
 & instruct & \textbf{22.96} & \textbf{49.88} & \textbf{8.61} \\  

 \midrule
\multirow{2}{*}{\textbf{EN-ZH}} & base & 7.23 & 10.62 & -52.06 \\
 & instruct & \textbf{38.73} & \textbf{32.53} & \textbf{17.38} \\ \bottomrule
\end{tabular}
\caption{Evaluation of the Qwen2-Audio-7B (base) and Qwen2-Audio-7B-Instruct (instruct) models before fine-tuning. The ``instruct'' model outperforms the ``base'' model for both English-to-German (EN-DE) and English-to-Chinese (EN-ZH) speech translation.}
\label{tab:study-base}
\end{table}

\newpage
\subsection{Encoder-Decoder Layer Pruning}
\label{sec:study-enc-dec-pruning}

The Qwen2-Audio is based on the encoder-decoder Transformer architecture \citep{Vaswani2017-attention}. It consists of an encoder for audio (\textit{audio\_tower}) and a decoder for text generation (\textit{language\_model}), each of which comprises 32 layers. We first experimented with layer pruning of both the encoder and decoder. However, inspired by the Distil-Whisper work \cite{Gandhi2023-Distil-Whisper}, we experimented with pruning decoder layers only, which achieved better results. In other words, pruning only the decoder from 32 layers to 24 layers outperformed pruning both the encoder and decoder into 24 layers.

In this experiment, we pruned 8 fixed middle layers, from the 12th to the 19th layer, inclusively. After fine-tuning both models with the English-to-German ACL 60/60 dataset, the model where only the decoder layers were pruned achieved 4+ additional points in terms of BLEU and chrF++ and 18+ points of COMET. It is worth noting that both the number of parameters and the storage footprint of this model is only 1\% larger than the model with both the encoder and decoder were pruned. Table~\ref{tab:study-enc-dec} shows the performance evaluation results of fine-tuning the English-to-German model after middle-layer pruning.

\begin{table*}[htbp]
\centering
\resizebox{\textwidth}{!}{%
\begin{tabular}{@{}lllllrrr@{}}
\toprule
\textbf{Language} & \textbf{Pruning} & \textbf{Metric} & \textbf{Pruned Layers} & \textbf{Model} & \textbf{BLEU} & \textbf{chrF/++} & \textbf{COMET} \\ \midrule
\multirow{6}{*}{\textbf{EN-DE}} 
    & \multirow{2}{*}{Middle}    & \multirow{2}{*}{n/a}   
    & \multirow{2}{*}{\small[12, 13, 14, 15, 16, 17, 18, 19]} 
    & Pruned       &  0.13 & 6.13 & -162.44  \\
    & & & & + FT & 30.81 & 58.45 & 31.95 \\
    \cmidrule{2-8}
    
    & \multirow{4}{*}{Iterative}    & \multirow{2}{*}{COMET}    
    & \multirow{2}{*}{\small[12, 1, 9, 15, 20, 27, 29, 5]}
    & Pruned       & 6.53	 & 31.15 & -41.96 \\
    & & & & + FT & 30.97 & 59.67 &	34.39 \\
    \cmidrule{3-8}
    
    &  & \multirow{2}{*}{chrF++}   
    & \multirow{2}{*}{\small[13, 3, 20, 9, 29, 1, 19, 27]} 
    & Pruned       & 10.78 & 39.58 & -44.40 \\
    & & & & + FT & \textbf{32.16} & \textbf{60.39} & \textbf{39.08} \\
    \midrule
    
\multirow{6}{*}{\textbf{EN-ZH}} 
    & \multirow{2}{*}{Middle}    & \multirow{2}{*}{n/a}   
    & \multirow{2}{*}{\small[12, 13, 14, 15, 16, 17, 18, 19]} 
    & Pruned       & 1.3 & 3.8 & -94.05  \\
    & & & & + FT & 45.84 & 40.48 & 40.37 \\
    \cmidrule{2-8}
    
    & \multirow{4}{*}{Iterative}    & \multirow{2}{*}{COMET}    
    & \multirow{2}{*}{\small[7, 9, 24, 3, 28, 6, 15, 18]}
    & Pruned       & 21.1 & 27.25 & 28.26  \\
    & & & & + FT & 51.85 & 46.36 & \textbf{58.64}  \\
    \cmidrule{3-8}
    
    &     & \multirow{2}{*}{chrF}   
    & \multirow{2}{*}{\small[7, 25, 3, 4, 29, 15, 26, 20]} 
    & Pruned       & 42.52 & 36.92 & 39.42  \\
    & & & & + FT & \textbf{53.05} & \textbf{47.23} & \uline{56.72}  \\

 \bottomrule
\end{tabular}
}
\caption{Comparison between \textbf{middle-layer pruning} and \textbf{iterative layer pruning}, with either COMET or chrF/chrF++ as the metric for measuring layer importance. Iterative layer pruning, i.e. removing layers one by one, and then evaluating the resulting model, outperforms middle-layer pruning. In particular, when chrF/chrF++ is used for layer importance evaluation, the resulting pruned model achieves better speech translation quality after fine-tuning on the ACL 60/60 dataset.}
\label{tab:study-iterative}
\end{table*}

\newpage
\subsection{Iterative vs Middle Layer Pruning}
\label{sec:study-middle-pruning}

In this section, we compare two common approaches to layer-wise pruning, namely iterative layer pruning and middle-layer pruning. Moreover, we compare using chrF/chrF++ and COMET for layer importance evaluation during iterative layer pruning. In all cases, we only work on decoder layers (cf.~Section \ref{sec:study-enc-dec-pruning}).

As discussed in Section \ref{sec:pruning}, we experimented with iterative pruning based on layer importance evaluation to identify and remove the layers that contribute least to translation quality. In contrast, in middle-layer pruning, we simply remove the 8 middle layers of the model, namely layers 12 through 19 out of the 32 decoder layers of the model.

Since the bottom layers of a model are closer to the input and the top layers are closer to the output, it is possible that both the top layers and the bottom layers are more important than the middle layers. In practice, the impact on model performance after pruning the middle layers varies across different models  \citep{Sajjad2023-EffectLayerDropping}.

In iterative layer pruning based on performance evaluation (using chrF++ for German and chrF for Chinese), the pruned layers are [1, 3, 9, 13, 19, 20, 27, 29] for German, and [3, 4, 7, 15, 20, 25, 26, 29] for Chinese. This shows a diverse layer selection that is not concentrated solely in the middle. As Table \ref{tab:study-iterative} illustrates, iterative layer pruning yields much better results than middle-layer pruning. For example, when pruning 8 middle layers of the English-to-Chinese model, the final evaluation scores of the resulting model are so low across all metrics (BLEU: 1.3, chrF: 3.8, COMET -94.05), compared to the scores achieved when the same number of layers is iteratively removed based on layer importance (BLEU: 42.52, chrF: 36.92, COMET 39.42).

Moreover, we experimented with both using chrF/chrF++ and COMET for evaluating pruned models during the iterative process. Interestingly, the final model obtained by using chrF/chrF++ for layer importance evaluation achieves better results (cf.~Table \ref{tab:study-iterative}). This might be due to the scientific nature of the ACL 60/60 dataset, and it requires future exploration with other datasets.

\setlength{\tabcolsep}{4pt} % Default value: 6pt

\begin{table*}[htbp]
\centering
\resizebox{\textwidth}{!}{%
\begin{tabular}{@{}lcrrrc@{\hskip0.5em}rrrc@{\hskip0.5em}|cc@{}}
\toprule
& \multicolumn{4}{c}{\textbf{EN-DE}} & \multicolumn{4}{c}{\textbf{EN-ZH}} & &
\textbf{Params} & \textbf{Storage} \\ 
\cmidrule(rl){3-5} \cmidrule(rl){7-9} \addlinespace[3pt]
\textbf{Model}  & \textbf{Layers} & \textbf{BLEU}  & \textbf{chrF++} & \textbf{COMET} & & \textbf{BLEU}  & \textbf{chrF}  & \textbf{COMET} & &
\multicolumn{1}{c}{(B)} & \multicolumn{1}{c}{(GB)}
\\ \midrule

Pruned [8]  & \multirow{2}{*}{24/32} & 10.78 & 39.58 & -44.4 & & 42.52 & 36.92 & 39.42 & & \multirow{2}{*}{6.78} & \multirow{2}{*}{13.55} \\

+ Fine-tuned  & & 32.16 & 60.39 & 39.08 & & 53.05 & 47.23 & 56.72 & &  \\ \midrule

Pruned [10] & \multirow{2}{*}{22/32} &  4.54 & 26.33 & -114.73   & & 16.51 & 23.02 & 5.89 & & \multirow{2}{*}{6.37} & \multirow{2}{*}{12.74} \\

+ Fine-tuned & & 30.90 & 59.80 & 33.74 && 51.97 & 47.19 & 59.57 & & \\ \midrule

Pruned [12] & \multirow{2}{*}{20/32} & 3.08 & 19.45 & -154.69    & & 5.97 & 10.14 & -62.50 & & \multirow{2}{*}{5.97} & \multirow{2}{*}{11.93} \\

+ Fine-tuned & & 31.05 & 58.68 & 30.03 && 52.43 & 47.03 & 56.15 & & \\ \midrule

Pruned [16] & \multirow{2}{*}{16/32} & 0.06 & 10.44 & -182.25    & & 1.99 & 4.26 & -124.75 & & \multirow{2}{*}{5.16} & \multirow{2}{*}{10.32} \\

+ Fine-tuned & & 22.15 & 50.90 & -6.92 && 47.33 & 42.13 & 42.51 & & \\

\bottomrule
\end{tabular}
}
\caption{Comparison of iterative \textbf{pruning of 8 decoder layers} (which is the foundation of our pruning experiments) against \textbf{pruning 10, 12, and 16 decoder layers}. We observe that the quality of German degrades much more rapidly than that of Chinese, after pruning more than 8 layers. Moreover, pruning 16 layers from the Chinese model results in notably worse performance compared to pruning only 8, 10, or 12 layers, even after fine-tuning. When iteratively removing 16 layers from the German model, the removed layers are: [13, 3, 20, 9, 29, 1, 19, 27, 7, 26, 15, 18, 10, 17, 14, 30]. For the Chinese model, the removed layers are [7, 25, 3, 4, 29, 15, 26, 20, 24, 5, 10, 9, 27, 28, 18, 13]. The layers are listed from most to least important (left to right), according to the layer importance evaluation used in our iterative pruning method. In this experiment, all 32 encoder layers are kept intact; therefore, the table reports the remaining decoder layers as 24/32, 22/32, 20/32, and 16/32 encoder/decoder layers.
}
\label{tab:study-more-layers}
\end{table*}

% Layers for German: [13, 3, 20, 9, 29, 1, 19, 27, 7, 26, 15, 18, 10, 17, 14, 30]
% Layers for Chinese: [7, 25, 3, 4, 29, 15, 26, 20, 24, 5, 10, 9, 27, 28, 18, 13]

\subsection{Immediate Recovery}
In our main experiments, we fine-tuned the pruned models only after completing the entire pruning process (cf.~Section \ref{sec:pruning}). In order to understand the effect of accompanying iterative pruning with iterative recovery \citep{Wibowo2025-IteRABRe}, we conducted an extra experiment where we immediately fine-tuned the model after each layer pruning iteration. In other words, when pruning 8 layers, the fine-tuning is performed 8 times, once after each pruning step. 

By the end of the process, the pruned layers for German are [13, 1, 3, 17, 12, 4, 14, 21], based on layer importance evaluation. These layers differ from those selected when pruning without immediate recovery (cf.~Section \ref{sec:study-middle-pruning}), since fine-tuning after each iteration changes the relative importance of the remaining layers.

\begin{table}[H]
    \centering
    \begin{tabular}{lrrr}
    \toprule
    \textbf{Model} & \textbf{BLEU}  & \textbf{chrF++} & \textbf{COMET} \\ 
    \midrule
    pruned  & 10.78 & 39.58 & -44.40 \\
    + FT [after]   & \textbf{32.16} & \textbf{60.39} & \textbf{39.08} \\ \midrule
    pruned  & 30.96 & 59.02 & 30.07 \\
    {\scriptsize\faRedo} FT [immediate]    & 31.67 & 59.76 & 36.13 \\
    \bottomrule
    \end{tabular}
    \caption{Comparison of fine-tuning after the end of the pruning process (+ FT [after]) against immediate fine-tuning after each pruning iteration ({\scriptsize\faRedo} FT [immediate]). Fine-tuning only after completing the pruning process achieves better final performance. The results are for iterative pruning of 8 decoder layers from the English-to-German model.}
    \label{tab:study-recovery}
\end{table}

While this immediate recovery approach improved evaluation scores of the pruned model, it did not achieve performance gains over our standard method of fine-tuning only once after complete pruning. This might be due to overfitting caused by fine-tuning several times on the small in-domain dataset. Moreover, immediate recovery through fine-tuning after each pruning iteration is much more computationally intensive.

\subsection{Pruning more layers}
\label{sec:study-more-layers}

In our main pruning experiments (cf.~Section \ref{sec:pruning}), we pruned only 8 layers based on layer importance evaluation. We decided to experiment with pruning more layers to explore the level to which a model can be pruned while keeping a similar level of translation performance. As Table \ref{tab:study-more-layers} illustrates, we compare pruning 8, 10, 12, and 16 layers. We observe that the quality after pruning up to 12 layers and fine-tuning the pruned model on the ACL 60/60 dataset is close to pruning 8 layers. However, when pruning 16 layers, the quality starts to degrade considerably. It is worth noting that the results reported in Table \ref{tab:study-more-layers} are only for the pruning and initial fine-tuning steps, while Section \ref{sec:pruning} describes the whole process that involves knowledge distillation, extra compression with quantization, and further fine-tuning with QLoRA.

Pruning reduces storage footprint while accelerating inference speed by approximately 20\% and 40\% when 8 and 16 layers are pruned, respectively, compared to the unpruned baseline, after full-parameter fine-tuning of both models on the English-to-German in-domain data. In contrast, 4-bit quantization as implemented in QLoRA reduces storage at the cost of inference speed. Given that the shared task prioritized minimizing storage requirements, we applied QLoRA to the pruned model in the final fine-tuning stage. For deployment scenarios where inference speed is critical, however, the pruned model can be fine-tuned with standard LoRA instead of QLoRA to avoid quantization overhead.

\subsection{Fine-tuning before/after pruning}
\label{sec:study-fine-tuning}

It is common to fine-tune pruned models \textit{after} pruning \citep{Kim2023-SOLAR10}. As we pruned 1/4 of the decoder layers, obtaining valid translations for German was not possible without further training (cf.~Table \ref{tab:layer-pruning}). Similarly, fine-tuning the teacher model on in-domain data \mbox{\textit{before}} pruning is especially recommended for downstream tasks \citep{Li2020-TrainLargeCompress}.

\subsection{Out-of-Domain Data Size}
\label{sec:study-data-size}

After layer pruning, we added more data for fine-tuning the pruned model to recover the translation quality of the teacher model. We mixed the in-domain data (ACL 60/60), knowledge distillation data, and out-of-domain data from the CoVoST2 dataset. We experimented with different sizes of out-of-domain data, namely 10k, 50k, 80k and 100k randomly sampled segments. 

\begin{table}[H]
    \centering
    \begin{tabular}{lrrr}
    \toprule
    \textbf{Data Size} & \textbf{BLEU}  & \textbf{chrF++} & \textbf{COMET} \\ 
    \midrule
    10k  & 37.83 & 63.84 & 48.90 \\
    50k  & 38.47 & 64.09 & 54.56 \\
    80k  & \textbf{40.24} & 65.04 & 54.75 \\
    100k & 39.59 & \textbf{65.14} & \textbf{59.21} \\
    \bottomrule
    \end{tabular}
    \caption{Investigating the effect of the data sizes used from the out-of-domain CoVoST2 dataset. Increasing out-of-domain data from 10k to 50k or 80k improves the translation quality on the English-to-German test split of the ACL 60/60 dataset. When increasing the out-of-domain data from 80k to 100k, this only improves the COMET score, but not the BLEU and chrF++ scores.}
    \label{tab:study-covost-size}
\end{table}

As Table \ref{tab:study-covost-size} shows, for English-to-German translation, increasing the data size from 10k to 50k and 80k segments improves the overall performance. However, increasing the data size to 100k has diminishing returns, especially in terms of BLEU and chrF++ scores, while it improves the COMET score.

\section{Inference and Evaluation}

For inference, we use greedy generation by disabling the sampling options. We apply the prompt illustrated in Section \ref{sec:study-base}. We use a batch size of 1, and set the generation max length to 1024 tokens.

To evaluate our systems, we calculated BLEU \citep{Papineni2002-BLEU},  chrF++ \citep{Popovic2017-chrF++}, as implemented in the sacreBLEU library\footnote{\url{https://github.com/mjpost/sacrebleu}} \citep{Post2018-sacreBLEU}. While we use chrF++ for German, we use raw chrF for Chinese, following the author of the metric who noted that “the concept of \textit{Chinese words} is generally not clear” \citep{Popovic2017-chrF++}.  For semantic evaluation, we use COMET \citep{Rei2020-COMET}.\footnote{In particular, we used the \textit{“wmt20-comet-da”} model.} Table \ref{tab:qlora-kd} and Table \ref{tab:layer-pruning} report the results of the main experiments. Moreover, we conduct an ablation study (cf.~Section \ref{sec:ablation-study}) to investigate the effect of modifying some aspects of our experiments, such as the baseline model, the pruning approach, the number of pruned layers, and the data size.

\section{Conclusions and Future Work}

In this work, we showed that combining multiple compression techniques enables substantial model size reduction with minimal impact on speech translation performance. To conclude, QLoRA fine-tuning with knowledge distillation achieved superior translation quality compared to layer pruning alone, though with reduced model compression. To achieve higher compression ratios while preserving translation quality, we employed a combined approach using iterative layer pruning, quantization, knowledge distillation, and multi-stage fine-tuning. The code of our experiments is publicly available.\footnote{\url{https://github.com/ymoslem/Model-Compression}}

In future work, we plan to explore adaptive compression strategies that dynamically adjust pruning levels and quantization precision based on real-time deployment constraints such as memory limits and latency requirements.
Additionally, we aim to evaluate our compression techniques across more diverse datasets, including both authentic and synthetic training data, to better understand the generalization capabilities of our approach.
Given that Qwen2-Audio-Instruct relies on text prompts for generation, it would be interesting to investigate retrieval-augmented generation with few-shot prompting to improve the translation quality of compressed models.

\newpage
\bibliography{paperpile}

\end{document}